\documentclass[letterpaper]{article} 
\usepackage{aaai2026}  
\usepackage{times}  
\usepackage{helvet}  
\usepackage{courier}  
\usepackage[hyphens]{url}  
\usepackage{graphicx} 
\usepackage{natbib}  
\usepackage{caption} 
\usepackage{float}
\usepackage{xcolor}

\usepackage{booktabs}
\usepackage{amssymb}
\usepackage{amsmath}

\title{SG-XDEAT: Sparsity-Guided Cross-Dimensional and Cross-Encoding Attention with Target-Aware Conditioning in Tabular Learning}
\author {
    Chih-Chuan Cheng\textsuperscript{},
    Yi-Ju Tseng\textsuperscript{}
}
\affiliations {
    willie051955.cs12@nycu.edu.tw, yijutseng@cs.nycu.edu.tw
}

\begin{document}

\maketitle

\begin{abstract}
We propose SG-XDEAT (Sparsity-Guided Cross-Dimensional and Cross-Encoding Attention with Target-Aware Conditioning), a novel framework designed for supervised learning on tabular data. At its core, SG-XDEAT employs a dual-stream encoder that decomposes each input feature into two parallel representations: a raw value stream and a target-conditioned (label-aware) stream. These dual representations are then propagated through a hierarchical stack of attention-based modules. SG-XDEAT integrates three key components: (i) cross-dimensional self-attention, which captures intra-view dependencies among features within each stream; (ii) cross-encoding self-attention, which enables bidirectional interaction between raw and target-aware representations; and (iii) an Adaptive Sparse Self-Attention (ASSA) mechanism, which dynamically suppresses low-utility tokens by driving their attention weights toward zero—thereby mitigating the impact of noise. Empirical results on multiple public benchmarks show consistent gains over strong baselines, confirming that jointly modeling raw and target-aware views—while adaptively filtering noise—yields a more robust deep tabular learner.
\end{abstract}

\section{Introduction}
Tabular data plays a central role in numerous real-world applications that span domains such as medicine, finance, and transportation~\cite{shwartz2022tabular, borisov2022deep, somvanshi2024survey, ye2024closer}. Despite its prevalence, learning from tabular data poses significant challenges for deep learning models due to its lack of spatial or sequential structure and the presence of heterogeneous feature types~\cite{somvanshi2024survey}. Consequently, Gradient-Boosted Decision Trees (GBDTs) have long remained the dominant choice for tabular tasks~\cite{shwartz2022tabular}. However, recent years have seen growing interest in deep learning techniques in this domain~\cite{hwang2023recent}. Attention-based models like FT-Transformer~\cite{gorishniy2021revisiting} capture feature interactions effectively. Graph-based methods such as GANDALF~\cite{joseph2022gandalf} exploit relational structure but often struggle to scale. Meanwhile, language-model approaches like PTab~\cite{liu2022ptab} and TabLLM~\cite{hegselmann2023tabllm} convert tabular data into text so they can leverage pretrained language models, delivering strong performance at a high computational cost~\cite{somvanshi2024survey}.

Recent research has explored the potential benefits of incorporating label information into feature encoding to improve model performance in tabular data~\cite{jiang2025representation}. For categorical variables, label-aware techniques, such as target encoding~\cite{micci2001preprocessing} and Generalized Linear Mixed Model (GLMM) ~\cite{stroup2024generalized}, outperform one-hot, hashing, and ordinal encoding schemes, particularly when classes are imbalanced or categories are high in cardinality~\cite{pargent2022regularized}. For numerical variables, methods like Piecewise Linear Encoding (PLE)~\cite{gorishniy2022embeddings} split the value range using label-guided thresholds and learn an embedding for each segment, yielding substantial gains for multilayer perceptrons (MLP)- and Transformer-based architectures. Overall, label-conditioned encodings expose informative structure that unsupervised approaches miss.

Deep learning models for tabular data are particularly vulnerable to irrelevant or weakly informative features~\cite{mcelfresh2023neural}. Early efforts to address this issue relied primarily on data-driven filtering techniques, such as ranking features by mutual information or correlation and removing the least informative~\cite{li2017feature}. However, more recent research has shifted toward model-driven approaches that embed sparsity directly into the network architecture. Instead of preprocessing inputs to filter out noise, these methods enable the model to learn which features to disregard during training~\cite{arik2021tabnet, margeloiu2023weight}. Additionally, some approaches modify activation functions themselves, for example, replacing softmax with rectified-linear variants such as squared ReLU~\cite{zhou2024adapt} or leaky ReLU~\cite{fiedler2021simple}, to further enhance robustness against noisy inputs. Overall, this transition from input filtering to architectural adaptation reflects a growing emphasis on making deep learning models inherently resilient to non-informative features.

To address the challenges mentioned above, namely susceptibility to irrelevant features and limited ability to leverage label information, we propose SG-XDEAT, a model that combines supervised feature representations with architectural mechanisms for noise suppression. SG-XDEAT adopts a dual-stream architecture that separately encodes raw inputs and label-aware representations to leverage label information. It incorporates attention modules to capture both cross-feature and cross-encoding dependencies. Furthermore, a hybrid sparse attention mechanism is introduced to dynamically downweight uninformative features.

We summarize the contributions of our paper as follows.
\begin{itemize}
\item \textbf{Target-Aware Conditioning} We introduce a tokenization stream with label-guided encodings: PLE for numerical features~\cite{gorishniy2022embeddings} and tree-based encoding~\cite{niculescu2009winning} for categorical ones.

\item \textbf{Dual-Path Transformer} We introduce two parallel attention streams: one captures cross-feature interactions, while the other learn cross-encoding dependencies.

\item \textbf{Adaptive Sparse Self-Attention} We refer to a hybrid attention module that combines softmax-based branch and squared-ReLU-based sparse branch~\cite{zhou2024adapt} to suppress noisy features while maintaining global context.

\item \textbf{Comprehensive Evaluation} We also verify the effectiveness and efficiency of SG-XDEAT, effectively bridging the gap between deep models and GBDTs.
\end{itemize}

\section{Related Work}
\textbf{Traditional Methods} 
Gradient Boosting Decision Trees (GBDT), including XGBoost~\cite{chen2016xgboost}, LightGBM~\cite{ke2017lightgbm}, and CatBoost~\cite{prokhorenkova2018catboost}, are widely used for tabular data due to their ability to handle mixed feature types, missing values, and non-linear patterns. These models offer strong predictive performance and remain the default choice in many structured data applications. In particular, CatBoost~\cite{prokhorenkova2018catboost} demonstrates state-of-the-art performance across various benchmark datasets, consistently outperforming both XGBoost~\cite{chen2016xgboost} and LightGBM~\cite{ke2017lightgbm} in terms of accuracy and stability. Despite their impressive performance, GBDT models are prone to overfitting, where deep trees tend to memorize the training data, including noise and irrelevant features, leading to poor generalization on unseen data.~\cite{costa2023recent}

\textbf{Deep Learning Models} 
Early work on deep learning for structured data relied on MLPs, yet these simple feed-forward networks rarely matched the performance of GBDT. Researchers then imported design principles from computer vision: residual MLPs in the style of ResNet, when carefully tuned, proved unexpectedly competitive~\cite{gorishniy2021revisiting}. Attention-based architectures soon followed—AutoInt~\cite{song2019autoint} replaces hand-crafted feature crosses with multi-head self-attention capable of capturing high-order interactions, while DCN-V2~\cite{wang2021dcn} augments explicit low-rank cross layers to satisfy the latency and scale requirements of industrial ranking systems. A parallel line of research grafts tree logic into neural networks: Neural Oblivious Decision Ensembles (NODE)~\cite{popov2019neural} integrate ensembles of oblivious decision trees within a fully differentiable scaffold, closing much of the gap to GBDTs while preserving interpretability. The latest advance, FT-Transformer~\cite{gorishniy2021revisiting}, pairs a feature tokenizer with a pre-norm Transformer encoder and has become the standard backbone in tabular learning.

\textbf{Target Aware Encoding}
Despite a variety of encoding strategies, traditional methods such as one-hot and ordinal encoding~\cite{bird2014categorical} do not incorporate any supervision from the target label. To address this, recent work has introduced label-aware encodings, particularly for categorical features~\cite{stroup2024generalized, larionov2020sampling, zeng2014necessary}. For example, regularized target encoders like the M-estimate and S-shrink variants~\cite{micci2001preprocessing} assign each category a smoothed estimate of the target mean, while DecisionTreeEncoder~\cite{niculescu2009winning} maps inputs to class probabilities based on leaf nodes of shallow trees. Similar ideas extend to numerical features: Piecewise Linear Encoding with Target guidance (PLE-T)~\cite{gorishniy2022embeddings} uses supervised splits to construct interval-based embeddings aligned with label distribution. 

However, these methods treat raw and target-aware representations as separate alternatives and do not model their interaction. In contrast, our approach introduces a dual-stream architecture that processes both representations in parallel and captures dependencies between them.

\textbf{Noise-Robust for Deep Learning Models}
Recent work has increasingly focused on integrating noise suppression directly into the model architecture. For example, ExcelFormer~\cite{chen2023excelformer} introduces a semi-permeable attention mask, built from pre-computed feature importance, that stops weak columns from sending information to stronger ones. The Leaky Gate applies a per-feature linear layer followed by a Leaky ReLU, down-weighting low-value inputs on the fly while making each feature’s impact transparent~\cite{fiedler2021simple}. In computer vision, Adaptive Sparse Self-Attention~\cite{zhou2024adapt} adds a parallel squared-ReLU branch that zeros small scores while a standard softmax branch preserves global context. 

Deep tabular models are often susceptible to irrelevant or weakly informative features~\cite{mcelfresh2023neural}, which can undermine both robustness and performance. To address this issue, we integrate the Adaptive Sparse Self-Attention~\cite{zhou2024adapt}, originally developed for vision tasks. By combining the squared-ReLU and softmax branches, our model learns to suppress low-utility attention scores, reducing the influence of noisy inputs without the need for manual feature selection.

\textbf{Dual-Path Transformer}
In multivariate time-series forecasting, it is increasingly common to divide attention along separate time and variable axes. For example, CrossFormer~\cite{zhang2023crossformer} embeds each series as a time-by-variable grid and applies a Two-Stage Attention (TSA) block: Stage 1 attends along time, Stage 2 along variables, capturing cross-time and cross-dimension patterns. TimeXer~\cite{wang2024timexer} builds on a similar idea, using self-attention within variable axes and cross-attention to fuse information from variable and time axes. 

In our tabular setup, each feature has two parallel representations after target-aware encoding: the raw value and its label-guided version. These two representations carry different types of signals, and capturing their dependencies is not straightforward. We discard a design such as CrossFormer~\cite{zhang2023crossformer}, which applies attention in a fixed order. Instead, we use a dual-path structure: one path attends across features within a view (i.e., within the raw value or within the label-guided version), while the other models cross-view interactions. This approach avoids arbitrary ordering and allows the model to better integrate the information from both representations.

\section{Methodology}
\begin{figure*}[t]
\centering
\includegraphics[width=1.0\textwidth]{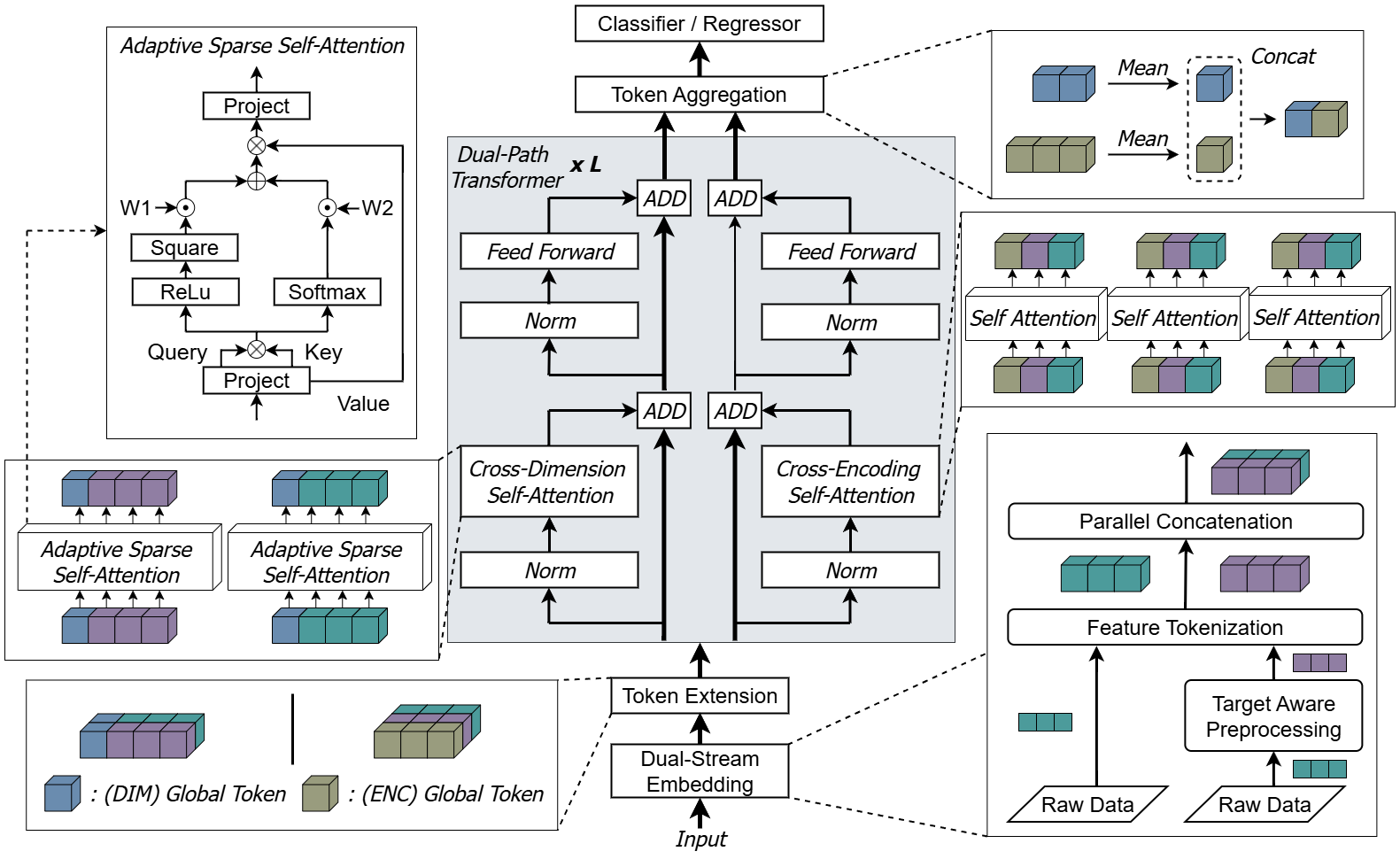}
\caption{The overview of SG-XDEAT. L is the layer number.}
\label{fig1}
\end{figure*}




Figure \ref{fig1} presents the overall architecture of \textbf{SG-XDEAT}. The raw input features are first processed by the \textbf{Dual-Stream Embedding} block, which generates two parallel streams: a \emph{raw stream} that retains the original feature values, and a \emph{target-aware stream} that encodes label-informed representations. Next, the \textbf{Token Extension} step adds global learnable tokens to both streams, enabling the model to incorporate instance-level contextual information. These enriched token streams are then passed through a \textbf{Dual-Path Transformer} stack. In each layer, (i) \emph{Cross-Encoding Self-Attention} aligns the raw and target-aware views of each feature, while (ii) \emph{Cross-Dimension Self-Attention}, equipped with an \textbf{Adaptive Sparse Self-Attention} module, facilitates information exchange across features while filtering out less informative tokens. After the transformer layers, the global tokens from both streams are mean-pooled and merged via the \textbf{Token Aggregation} module before being passed to the final classifier. Collectively, these components integrate raw and label-guided information, reduce noise, and produce a rich representation for tabular classification or regression. The following sections describe each key module in detail.

\subsection{Dual-Stream Embedding}

Each input sample is processed through two parallel embedding pathways. The \emph{raw stream} \(\mathcal{R}\) preserves the original values of all features, while the \emph{target-aware stream} \(\mathcal{T}\) transforms features based on the label information.

In the target-aware stream, categorical features are encoded using a \textbf{DecisionTreeEncoder}~\cite{niculescu2009winning}: a shallow decision tree is trained for each feature using the target label, and the class probability at the leaf node replaces the raw category. For numerical features, we use \textbf{PLE-T}~\cite{gorishniy2022embeddings}, which fits a single-feature decision tree to the target. The resulting leaf bins define discrete intervals, and the corresponding bin index is mapped to a learnable embedding vector. In the raw stream, feature values are preserved without modification.

Both streams are then passed through a tokenizer \(E(\cdot)\), which projects each feature into a \(d\)-dimensional embedding space. This yields two embedding matrices:

\[
\begin{aligned}
\mathbf{R} &= E(\mathcal{R}) \in \mathbb{R}^{F \times d} \quad \text{(Raw Embeddings)} \\
\mathbf{T} &= E(\mathcal{T}) \in \mathbb{R}^{F \times d} \quad \text{(Target-Aware Embeddings)} \\
\end{aligned}
\]

where \(F\) is the number of features and \(d\) is the embedding dimension. The raw and target-aware embeddings are then concatenated along a new stream dimension to form \(\mathbf{D}\), integrating both streams for Token Extension module:

\[
\begin{aligned}
\mathbf{D} &= \text{Concat}(\mathbf{T}, \mathbf{R}) \in \mathbb{R}^{2 \times F \times d}
\end{aligned}
\]

\subsection{Token Extension}
We introduce two types of learnable global tokens. The \emph{dimension-level tokens} \(\mathbf{g}_{\text{dim}} \in \mathbb{R}^{2 \times d}\) are prepended to the raw and target-aware representations to model cross-feature dependencies. The \emph{encoding-level tokens} \(\mathbf{g}_{\text{enc}} \in \mathbb{R}^{F \times d}\) capture interactions between the two encoding views. The resulting extended representations are:
\[
\begin{aligned}
\mathbf{D}_{\text{dim}} &= [\mathbf{g}_{\text{dim}}; \mathbf{D}] \in \mathbb{R}^{2 \times (F+1) \times d} \\
\mathbf{D}_{\text{enc}} &= [\mathbf{g}_{\text{enc}}; \mathbf{D}] \in \mathbb{R}^{3 \times F \times d}
\end{aligned}
\]

Here, \(\mathbf{D}_{\text{dim}}\) appends \(\mathbf{g}_{\text{dim}}\) along the feature axis, while \(\mathbf{D}_{\text{enc}}\) includes \(\mathbf{g}_{\text{enc}}\) as another token group. These augmentations enable models to capture both cross-feature patterns and cross-view relations.

\subsection{Dual-Path Transformer}
The Dual-Path Transformer comprises two attention modules—\textit{Cross-Encoding Self-Attention} and \textit{Cross-Dimension Self-Attention}—that extract complementary information from \(\mathbf{D}_{\text{dim}}\) and \(\mathbf{D}_{\text{enc}}\).

\subsubsection{Cross-Encoding Self-Attention}
This module operates on the encoding-level token group \(\mathbf{D}_{\text{enc}} \in \mathbb{R}^{3 \times F \times d}\), which includes global encoding tokens, raw feature tokens, and target-aware tokens. To enable feature-wise attention, the tensor is first reshaped to \(\mathbb{R}^{F \times 3 \times d}\) to treat each feature independently. A standard multi-head self-attention block is applied independently to each of the \(F\) features to model interactions among the three token types. For a given feature, the attention is computed as:
\[
\text{Attention}(Q, K, V) = \text{softmax}\left(\frac{QK^\top}{\sqrt{d}}\right)V,
\]
where the query, key, and value matrices are computed: \(Q = \mathbf{D}_{\text{enc}}W_Q\), \(K = \mathbf{D}_{\text{enc}}W_K\), and \(V = \mathbf{D}_{\text{enc}}W_V\), with \(W_Q, W_K, W_V \in \mathbb{R}^{d \times d}\) being learnable parameters.

This setup enables the model to learn encoding-level relations for each feature and use the global token to capture interactions between the raw and target-aware representations.

\subsubsection{Cross-Dimension Self-Attention}
This module builds on the \textit{Adaptive Sparse Self-Attention} mechanism, enabling the model to focus on informative features while filtering out less relevant ones. It operates on dimension-level tokens \(\mathbf{D}_{\text{dim}} \in \mathbb{R}^{2 \times (F+1) \times d}\), which consist of raw and target-aware token groups, each prepended with a global dimension token. To capture cross-feature dependencies, the raw and target-aware representations are processed with attention applied across the feature dimension within each group.

Two attention branches are then applied in parallel:

\begin{itemize}
    \item \textbf{Softmax Branch:} This branch applies scaled dot-product attention, capturing fine-grained dependencies:
    \[
    \mathbf{Emb}_{\text{soft}} = \text{softmax}\left(\frac{QK^\top}{\sqrt{d}}\right)V
    \]
    
    \item \textbf{Squared ReLU Branch:} To enhance sparsity and suppress noisy signals, this branch replaces the softmax with a squared ReLU activation:
    \[
    \mathbf{Emb}_{\text{relu}} = \left[\text{ReLU}\left(\frac{QK^\top}{\sqrt{d}}\right)\right]^2 V
    \]
\end{itemize}

Given the input tokens \(\mathbf{D}_{\text{dim}} \in \mathbb{R}^{(F+1) \times d}\) from one view, we compute the query, key, and value matrices:
\[
Q = \mathbf{D}_{\text{dim}} W_Q, \quad K = \mathbf{D}_{\text{dim}} W_K, \quad V = \mathbf{D}_{\text{dim}} W_V
\]
where \(W_Q, W_K, W_V \in \mathbb{R}^{d \times d}\) are learnable parameters.

The outputs from both branches are combined using learned weights. Let \(a_1, a_2 \in \mathbb{R}\) be scalar parameters (initialized to 1), and define the normalized weights \(w_1, w_2\) using a softmax over \((a_1, a_2)\):
\[
w_n = \frac{e^{a_n}}{e^{a_1} + e^{a_2}}, \quad n = \{1, 2\}
\]
The final attention output is computed as a weighted sum:
\[
\mathbf{D}_{\text{dim}} = w_1 \cdot \mathbf{Emb}_{\text{soft}} + w_2 \cdot \mathbf{Emb}_{\text{relu}}
\]
This approach enables the model to weigh standard attention against a denoising pathway, enhancing the relevance of feature interactions.

\subsection{Token Aggregation}
The model performs mean pooling independently over the global tokens, and the pooled vectors are then concatenated to form the final representation:
\[
\begin{aligned}
\mathbf{g}_{\text{dim}}^{\text{mean}} &= \text{Mean}(\mathbf{g}_{\text{dim}}) \in \mathbb{R}^{d}, \quad
\mathbf{g}_{\text{enc}}^{\text{mean}} = \text{Mean}(\mathbf{g}_{\text{enc}}) \in \mathbb{R}^{d}
\end{aligned}
\]
\[
\mathbf{z} = \text{Concat}(\mathbf{g}_{\text{dim}}^{\text{mean}}, \mathbf{g}_{\text{enc}}^{\text{mean}}) \in \mathbb{R}^{2d}
\]
This fused vector \(\mathbf{z}\) is subsequently passed to the classifier.

\subsection{Classifier / Regressor}
The aggregated vector \(\mathbf{z} \in \mathbb{R}^{2d}\) is passed to the classifier:
\[
\hat{y} = \text{Linear}(\text{ReLU}(\text{LayerNorm}(\mathbf{z}))).
\]

\section{Experiments}
\subsection{Experimental Setup}

\subsubsection{Dataset}
To evaluate the generalizability of our model, we use a set of benchmark datasets covering regression, binary classification, and multiclass classification tasks. The datasets vary in feature composition—some are only numerical, others entirely categorical, and some include both. The selected datasets are Gesture Phase (GE)~\cite{madeo2013gesture}, KDD Internet Usage (KD)~\cite{kehoe1996surveying}, Adult (AD)~\cite{kohavi1996scaling}, California Housing (CA)~\cite{pace1997sparse}, and Higgs Small (HI)~\cite{baldi2014searching}. Detailed statistics and configuration are provided in Table~\ref{table:dataset-stats} and Appendix A.

\begin{table*}[t]
\centering
\begin{tabular}{l|c|c|c|c|c|c|c|c}
\toprule
\textbf{Name} & \textbf{Abbr} & \textbf{\#Train} & \textbf{\#Val} & \textbf{\#Test} & \textbf{\#Num} & \textbf{\#Cat} & \textbf{Task Type} & \textbf{Batch Size} \\
\midrule
Gesture Phase      & GE  & 6,318  & 1,580  & 1,975  & 32 & 0   & Multiclass   & 128 \\
KDD Internet Usage & KD  & 6,468  & 1,618  & 2,022  & 0  & 68  & Binclass     & 128 \\
Adult              & AD  & 26,048 & 6,513  & 16,281 & 6  & 8   & Binclass     & 256 \\
California Housing & CA  & 13,209 & 3,303  & 4,128  & 8  & 0   & Regression   & 256 \\
Higgs Small        & HI  & 62,752 & 15,688 & 19,610 & 28 & 0   & Binclass     & 512 \\
\bottomrule
\end{tabular}
\caption{Dataset statistics}
\label{table:dataset-stats}
\end{table*}

\subsubsection{Baseline}

\begin{table*}[t]
\centering
\begin{tabular}{l|ccccc|c}
\toprule
\textbf{Datasets} 
& \textbf{CA} 
& \textbf{GE} 
& \textbf{AD} 
& \textbf{KD} 
& \textbf{HI} 
& -- \\
\cmidrule(lr){2-2} \cmidrule(lr){3-6} \cmidrule(lr){7-7} 
\textbf{Metrics} 
& \multicolumn{1}{c}{RMSE\,$\downarrow$} 
& \multicolumn{4}{c}{Accuracy\,$\uparrow$} 
& \multicolumn{1}{c}{Rank $\pm$ std} \\
\midrule
XGBoost        & \textbf{0.451 (1.0)} & \textbf{0.685 (1.0)} & \underline{0.871 (2.0)} & 0.902 (3.0) & 0.727 (4.0) & \underline{2.2 ± 1.17} \\
\midrule
MLP            & 0.499 (7.0) & 0.651 (5.0) & 0.858 (6.0) & 0.892 (7.0) & 0.725 (7.0) & 6.4 ± 0.80 \\
Resnet         & 0.489 (5.0) & 0.657 (4.0) & 0.852 (7.0) & 0.894 (6.0) & \textbf{0.734 (1.0)} & 4.6 ± 2.06 \\
DCN-V2         & 0.488 (4.0) & 0.634 (6.0) & 0.859 (4.5) & 0.899 (4.0) & 0.726 (5.5) & 4.8 ± 0.81 \\
AutoInt        & 0.490 (6.0) & 0.602 (7.0) & 0.859 (4.5) & 0.898 (5.0) & 0.726 (5.5) & 5.6 ± 0.86 \\
FT-Transformer & 0.472 (3.0) & \underline{0.677 (2.0)} & 0.861 (3.0) & \textbf{0.903 (1.5)} & \underline{0.732 (2.5)} & 2.4 ± 0.58 \\
\midrule
SG-XDEAT       & \underline{0.454 (2.0)} & 0.675 (3.0) & \textbf{0.872 (1.0)} & \textbf{0.903 (1.5)} & \underline{0.732 (2.5)} & \textbf{2.0 ± 0.71} \\
\bottomrule
\end{tabular}
\caption{Comparison of performance across various benchmark models. Numbers in parentheses denote ranks of performance, and the \textbf{best} and \underline{second-best} results are highlighted.}
\label{table:main_results}
\end{table*}

We evaluated our approach against a range of baselines commonly used for tabular data. Traditional methods include gradient boosting models such as \textbf{XGBoost}~\cite{chen2016xgboost}. Deep learning baselines include a standard \textbf{MLP}~\cite{gorishniy2021revisiting} and several advanced architectures: \textbf{ResNet}~\cite{gorishniy2021revisiting}, \textbf{DCN-V2}~\cite{wang2021dcn}, \textbf{AutoInt}~\cite{song2019autoint}, and \textbf{FT-Transformer}~\cite{gorishniy2021revisiting}. This selection ensures a comprehensive comparison across both conventional and modern learning paradigms.

\subsubsection{Implementation Details}
All experiments were implemented in Python 3.10 using PyTorch 2.5.1 and executed on an NVIDIA RTX 4090 GPU with 24GB of memory.

All deep learning models were evaluated using the same preprocessing: quantile transformation for input features (via Scikit-learn). In addition, regression targets were standardized across all methods.

All models were trained using the AdamW optimizer, along with a cosine annealing scheduler. A linear warm-up strategy was applied during the first 10 epochs, and each model was trained for a maximum of 200 epochs. Early stopping was employed with a patience of 15 epochs based on validation performance.

The hyperparameters of the models were tuned individually for each dataset using a multi-objective optimization strategy based on the Optuna framework~\cite{akiba2019optuna}. The optimization process aimed to balance performance on both the training and validation datasets to ensure good generalization. To promote reproducibility and consistent results, we set the random seed to 42 during the search process. The search ranges and grids used for hyperparameter tuning are provided in Appendix B.

For classification tasks, accuracy was used as the evaluation metric, while for regression tasks, root mean squared error (RMSE) was used. Once the optimal configuration was found, each model was retrained 15 times (using random seeds from 0 to 14), and the final results were reported as the average over these runs.

\subsection{Experimental Results}

\begin{figure*}[t]
\centering
\includegraphics[width=0.8\textwidth]{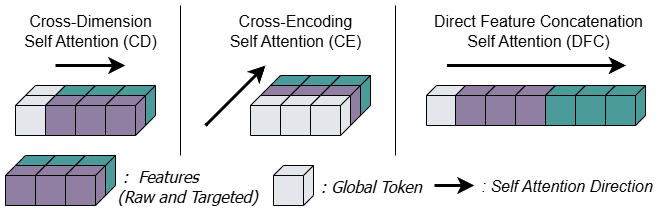}
\caption{Illustration of architectural variants used in the ablation study.}
\label{fig2}
\end{figure*}

\subsubsection{Performance Comparison}
Table~\ref{table:main_results} presents a comprehensive comparison of benchmark models across five tabular datasets. The proposed model, SG-XDEAT, achieves state-of-the-art performance on the Adult (AD) dataset with an accuracy of 0.872, and obtains the second-best RMSE on the California Housing (CA) dataset (0.454)—a result very close to the best-performing XGBoost (0.451), while substantially outperforming all other deep learning architectures. It also demonstrates competitive results on the remaining datasets, including a joint-best accuracy on the KDD Internet Usage (KD) dataset (0.903). These results are consistent with the findings reported in the FT-Transformer research, where XGBoost is shown to perform strongly on certain datasets, and FT-Transformer also exhibits competitive performance. Among all deep learning-based methods, SG-XDEAT clearly stands out in most datasets and achieves the best average rank (2.0 ± 0.71), underscoring its robustness and strong generalizability. More detailed, one-sided Wilcoxon Test to statistically confirm these findings and further evaluate the significance of differences between SG-XDEAT and other models will be provided in Appendix C.

\subsubsection{Ablation Analysis of Architectural Components}
\begin{table}[t]
\centering
\begin{tabular}{l|ccccc}
\toprule
\textbf{Datasets} 
& \textbf{CA} 
& \textbf{GE} 
& \textbf{AD} 
& \textbf{KD} 
& \textbf{HI} \\
\cmidrule(lr){2-2} \cmidrule(lr){3-6} 
\textbf{Metrics} 
& \multicolumn{1}{c}{RMSE\,$\downarrow$} 
& \multicolumn{4}{c}{Accuracy\,$\uparrow$} \\
\midrule
DFC    & 0.480 & 0.649 & 0.869 & 0.902 & \textbf{0.732} \\
\midrule
CD               & 0.463 & \textbf{0.678} & 0.871 & \textbf{0.903} & \textbf{0.732} \\
CE               & 0.459 & 0.626 & \textbf{0.872} & 0.899 & 0.727 \\
\midrule
CD + CE   & \textbf{0.454} & 0.675 & \textbf{0.872} & \textbf{0.903} & \textbf{0.732} \\
\bottomrule
\end{tabular}
\caption{Ablation results for architectural components (visualized in Figure~\ref{fig2}). \textbf{CD} = Cross-Dimension, \textbf{CE} = Cross-Encoding, \textbf{DFC} = Direct Feature Concatenation. \textbf{Best} results are highlighted.}
\label{table:ablation_study}
\end{table}

To evaluate the contributions of the proposed architectural modules, we conducted an ablation study comparing three variants of our model: Cross-Dimension Self Attention (CD), Cross-Encoding Self Attention (CE), and their combination (CD+CE). Additionally, we include Direct Feature Concatenation (DFC) as a baseline strategy, which represents the most common approach—applying standard self-attention to concatenated feature representations. It is important to note that Cross-Dimension Self Attention (CD) in this ablation does not incorporate the adaptive sparsity mechanism used in the SG-XDEAT model. The architectural designs of these variants are illustrated in Figure~\ref{fig2}.

As shown in Table~\ref{table:ablation_study}, the combined configuration CD+CE consistently achieves the best performance across most datasets, including the lowest RMSE on CA (0.454) and the highest accuracy on AD (0.872), KD (0.903), and HI (0.732), demonstrating the complementary nature of the two mechanisms. Individually, both CD and CE contribute to performance improvements: CD, which focuses on feature-level fusion, is especially effective on GE, KD, and HI, while CE—which emphasizes encoding-level fusion—achieves the highest accuracy on AD and also performs best on CA. These observations indicate that different datasets benefit from different forms of interaction modeling, and thus, jointly considering both feature-level and encoding-level fusion is crucial for achieving robust and generalizable performance across diverse tabular tasks.

Furthermore, the baseline DFC configuration, which lacks explicit modeling of both feature-wise and encoding-wise dependencies, performs worse than CD+CE across most datasets—particularly on CA and GE. These results demonstrate that both CD and CE play an important role in improving model performance, leading to a more effective and robust architecture overall.

\subsubsection{Analysis of Different Input Strategies}
To evaluate the impact of target-aware information on input representations, we compare two distinct streams: the raw stream, which uses the original input features, and the target-aware stream, which integrates target information. As shown in Table~\ref{table: embeddings}, the effectiveness of each stream varies across datasets. The raw stream performs better on GE and KD, suggesting that preserving the original feature semantics is advantageous for these cases. On the other hand, the target-aware stream outperforms on CA and AD, demonstrating that incorporating label-aware context can improve discriminative power. For the HI dataset, both streams show similar results.

These findings emphasize that different datasets respond uniquely to raw and target-aware representations. However, the \textbf{DFC} method, which directly combines the raw and target-aware streams, fails to capture the dependencies between them effectively. In contrast, \textbf{CD+CE} can capture both cross-view and cross-feature interactions and ultimately improve performance.

\begin{table}[t]
\centering
\begin{tabular}{l|ccccc}
\toprule
\textbf{Datasets} 
& \textbf{CA} 
& \textbf{GE} 
& \textbf{AD} 
& \textbf{KD} 
& \textbf{HI} \\
\cmidrule(lr){2-2} \cmidrule(lr){3-6} 
\textbf{Metrics} 
& \multicolumn{1}{c}{RMSE\,$\downarrow$} 
& \multicolumn{4}{c}{Accuracy\,$\uparrow$} \\
\midrule
Raw         & 0.483 & \underline{0.665} & 0.859 & \textbf{0.903} & \textbf{0.732} \\
Targeted    & \underline{0.479} & 0.655 & 0.863 & 0.888 & \textbf{0.732} \\
\midrule
DFC         & 0.480 & 0.649 & \underline{0.869} & 0.902 & \textbf{0.732} \\
\midrule
CD + CE    & \textbf{0.454} & \textbf{0.675} & \textbf{0.872} & \textbf{0.903} & \textbf{0.732} \\
\bottomrule
\end{tabular}
\caption{Performance comparison between different input strategies. The \textbf{Raw} setting uses original features only, while the \textbf{Targeted} variant incorporates label-dependent encoding. The \textbf{DFC} and \textbf{CD+CE} methods are visualized in Figure~\ref{fig2}. \textbf{Best} and \underline{second-best} results are highlighted.}
\label{table: embeddings}
\end{table}

\subsubsection{Effectiveness of Adaptive Sparse Self-Attention}

\begin{figure}[t]
\centering
\includegraphics[width=0.45\textwidth]{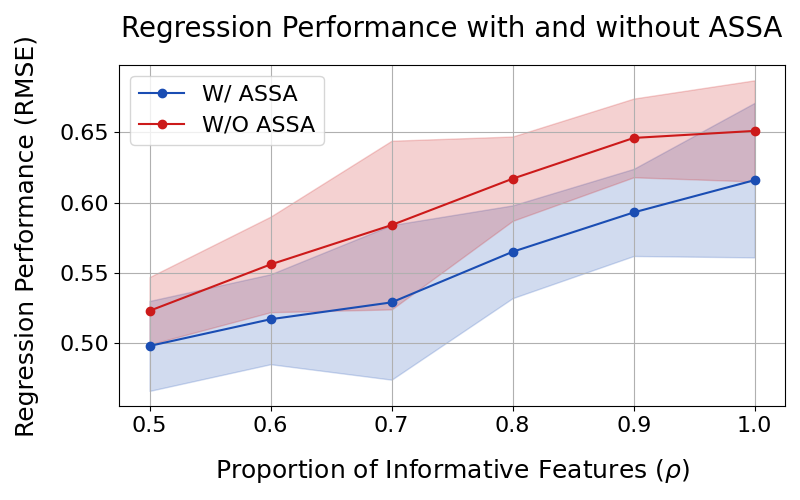}
\caption{Regression performance (RMSE) on synthetic datasets with and without ASSA. A lower $\rho$ implies a higher proportion of irrelevant (noisy) features. Lower RMSE indicates better predictive performance.}
\label{fig:ASSA}
\end{figure}


To assess the robustness of Adaptive Sparse Self-Attention (ASSA) under varying levels of feature redundancy, we construct a synthetic regression benchmark in which only a controlled subset of the input features is informative. Each input sample is represented as $x \in \mathbb{R}^d$ with $d = 100$, and a proportion $\rho \in \{0.5, 0.6, \dots, 1.0\}$ of the features contribute to the target. The number of informative dimensions is defined as $d_{\text{useful}} = \lfloor \rho \cdot d \rfloor$, and the informative subvector is given by $x_{\text{useful}} \in \mathbb{R}^{d_{\text{useful}}}$, consisting of the first $d_{\text{useful}}$ elements of $x$. The remaining $(1 - \rho) \cdot d$ features are uninformative and serve as distractors.  We focus on $\rho \geq 0.5$ to maintain a meaningful signal-to-noise ratio, enabling attention-based models to effectively exploit informative features and better reflect practical scenarios.

The target value $y \in \mathbb{R}$ is generated by applying a fixed, randomly initialized multi-layer perceptron (MLP) to the informative subvector:
\[
y = f_{\text{MLP}}(x_{\text{useful}}).
\]
Subsequently, standard normalization is applied:
\[
y \leftarrow \frac{y - \mu_y}{\sigma_y},
\]
where $\mu_y$ and $\sigma_y$ denote the mean and standard deviation of the target values computed from the training set.

This setup enables us to assess whether ASSA can suppress attention to irrelevant features and focus on informative ones. Details of the synthetic dataset generation and model architecture settings are provided in Appendix C.

Figure~\ref{fig:ASSA} illustrates the impact of ASSA across varying levels of feature redundancy. Across all settings of $\rho$, models equipped with ASSA consistently achieve lower RMSE compared to their counterparts without ASSA. This suggests that the adaptive sparsity mechanism enables the model to focus on relevant inputs even in the presence of substantial noise, leading to more robust and accurate predictions. Interestingly, even when \( \rho = 1 \), meaning that all features are informative, the model with ASSA still outperforms the model without it. This may be attributed to ASSA's ability to better prioritize and focus on the most relevant features, even in scenarios where all features contribute useful information. Therefore, ASSA's mechanism of enforcing sparse attention might still enhance model performance, even when there is no explicit need to filter out irrelevant features.

\section{Conclusion}
We introduced SG-XDEAT, a sparsity-guided attention framework designed for deep learning on tabular data. By capturing both cross-feature and cross-encoding dependencies through a dual-path attention design, SG-XDEAT effectively leverages raw and label-informed representations. The incorporation of an adaptive sparse attention mechanism further improves robustness by suppressing noisy or low-relevance signals. Empirical results across a range of benchmarks show that SG-XDEAT consistently outperforms strong baselines, helping close the gap between deep models and gradient-boosted decision trees. These results underscore the benefit of integrating label-aware encoding with structured attention for tabular prediction tasks.

\section*{Acknowledgments}
This research was supported by the National Science and Technology Council of Taiwan (grant nos. NSTC 114-2221-E-A49 -061) and the Higher Education Sprout Project of National Yang Ming Chiao Tung University and the Ministry of Education, Taiwan (grant no. CGMH-NYCU-114-CORPG2P0072). The funders had no role in the study design and procedures; data collection, management, analysis, and interpretation; manuscript preparation, review, and approval; or the decision to submit the manuscript for publication.

\bibliography{references}

\appendix
\section{Appendix A}
\subsection{Dataset Description}
The datasets differ in feature types: some are purely numerical, others entirely categorical, and some are mixed. The selected datasets include Gesture Phase (GE)~\cite{madeo2013gesture}, KDD Internet Usage (KD)~\cite{kehoe1996surveying}, Adult (AD)~\cite{kohavi1996scaling}, California Housing (CA)~\cite{pace1997sparse}, and Higgs Small (HI)~\cite{baldi2014searching}. Among these, AD, KD, HI, and GE are retrieved from the OpenML platform, while CA is accessed via the scikit-learn library.

\subsubsection{California Housing (Scikit-learn)}
The California Housing dataset~\cite{pace1997sparse} pertains to houses found in a given California district and some summary statistics on them based on 1990 census data. The final data contained 20,640 observations on 8 numerical variables. The dependent variable is median house value.

\subsubsection{Adult (OpenML ID=1590)}
The Adult dataset~\cite{kohavi1996scaling} is a benchmark dataset commonly used for classification tasks in machine learning. It contains 48,842 instances with 14 features, which include 8 categorical and 6 integer types. The goal is to predict whether an individual's annual income exceeds 50,000, based on census information such as age, education level, occupation, and marital status. 

\subsubsection{KDD Internet Usage (OpenML ID=981)}
The KDD Internet Usage dataset~\cite{kehoe1996surveying}, originally compiled by the Georgia Tech Research Corporation as part of the GVU's WWW User Surveys, provides detailed information on users’ internet usage patterns and demographic characteristics. The dataset contains 10,120 instances and consists exclusively of 68 categorical features. The objective of this task is to predict which users are likely to pay for internet access at work.

\subsubsection{HIGGS (OpenML ID=23512)}
The HIGGS Small dataset~\cite{baldi2014searching} is a physics dataset generated via Monte Carlo simulations, containing approximately 98,000 samples. It includes 28 numerical features, divided into 21 low-level and 7 high-level features. The low-level features represent kinematic properties directly measured by particle detectors. In contrast, the high-level features are derived from the low-level ones using domain knowledge, designed by physicists to enhance discrimination between signal and background events. The dataset is used for classifying particle collision events.

\subsubsection{Gesture (OpenML ID=4538)}
The dataset~\cite{madeo2013gesture} contains frame-based gesture phase segmentation data from 7 videos, each with approximately 1,400 to 2,700 frames. For each frame, two types of motion features are provided: 3D positions of the hands, wrists, head, and spine (from raw files, 18 features), and velocity/acceleration of the hands and wrists (from processed files, 32 features). Combined, each frame yields up to 50 numeric features with a class label, enabling per-frame gesture phase prediction.

\section{Appendix B}
\subsection{Model Hyperparameter Configuration (Optuna)}
For each model, we perform hyperparameter optimization using Optuna~\cite{akiba2019optuna}. The search space follows the settings used in tabular deep learning benchmarks~\cite{gorishniy2021revisiting}. 

\subsection{XGBoost}
\noindent\textbf{Implementation:} We use the official \texttt{xgboost} library~\cite{chen2016xgboost}. The following parameters are fixed throughout all experiments:
\begin{itemize}
\item \texttt{booster = "gbtree"}
\item \texttt{early\_stopping\_rounds = 50}
\item \texttt{n\_estimators = 2000}
\end{itemize}
The remaining hyperparameters are tuned according to the search space defined in Table~\ref{table:xgboost_hyperparam}.

\begin{table}[H]
\centering
\begin{tabular}{l|l}
\toprule
\textbf{Parameter} & \textbf{Distribution} \\
\midrule
Max depth               & UniformInt[3, 10] \\
Min child weight        & LogUniform[1e-8, 1e5] \\
Subsample               & Uniform[0.5, 1] \\
Learning rate           & LogUniform[1e-5, 1] \\
Col sample by level     & Uniform[0.5, 1] \\
Col sample by tree      & Uniform[0.5, 1] \\
Gamma                   & LogUniform[1e-8, 1e2] \\
Lambda                  & LogUniform[1e-8, 1e2] \\
Alpha                   & LogUniform[1e-8, 1e2] \\
\midrule
\# Iterations           & 100 \\
\bottomrule
\end{tabular}
\caption{XGBoost hyperparameter search space.}
\label{table:xgboost_hyperparam}
\end{table}

\subsection{Deep Learning Models}
\noindent\textbf{Implementation:}  
We implemented all deep learning models using the official codebase from Gorishniy et al.~\cite{gorishniy2021revisiting}, which serves as the foundation for many recent benchmarks on tabular data. The models include:

\begin{itemize}
    \item MLP (Table~\ref{table:mlp_hyperparam})~\cite{gorishniy2021revisiting}
    \item ResNet (Table~\ref{table:resnet_hyperparam})~\cite{gorishniy2021revisiting}
    \item DCN V2 (Table~\ref{table:dcnv2_hyperparam})~\cite{wang2021dcn}
    \item AutoInt (Table~\ref{table:autoint_hyperparam})~\cite{song2019autoint}
    \item FT-Transformer (Table~\ref{table:fttransformer_hyperparam})~\cite{gorishniy2021revisiting}
\end{itemize}

\noindent Each corresponding table presents the hyperparameter search space used for tuning these models. The hyperparameter configuration for our proposed method, SG-XDEAT, is provided separately in Table~\ref{table:SG-XDEAT_hyperparam}.

\begin{table}[H]
\centering
\begin{tabular}{l|l}
\toprule
\textbf{Parameter} & \textbf{Distribution} \\
\midrule
\# Layers                & UniformInt[1, 8] \\
Layer size               & UniformInt[1, 512] \\
Dropout                  & Uniform[0, 0.5] \\
Learning rate            & LogUniform[1e-5, 1e-2] \\
Weight decay             & LogUniform[1e-6, 1e-3] \\
Category embedding size  & UniformInt[64, 512] \\
\midrule
\# Iterations           & 100 \\
\bottomrule
\end{tabular}
\caption{MLP hyperparameter search space.}
\label{table:mlp_hyperparam}
\end{table}

\begin{table}[H]
\centering
\begin{tabular}{l|l}
\toprule
\textbf{Parameter} & \textbf{Distribution} \\
\midrule
\# Layers               & UniformInt[1, 8] \\
Layer size              & UniformInt[64, 512] \\
Hidden factor           & Uniform[1, 4] \\
Hidden dropout          & Uniform[0, 0.5] \\
Residual dropout        & Uniform[0, 0.5] \\
Learning rate           & LogUniform[1e-5, 1e-2] \\
Weight decay            & LogUniform[1e-6, 1e-3] \\
Category embedding size & UniformInt[64, 512] \\
\midrule
\# Iterations           & 100 \\
\bottomrule
\end{tabular}
\caption{ResNet hyperparameter search space.}
\label{table:resnet_hyperparam}
\end{table}

\begin{table}[H]
\centering
\begin{tabular}{l|l}
\toprule
\textbf{Parameter} & \textbf{Distribution} \\
\midrule
\# Cross layers           & UniformInt[1, 8] \\
\# Hidden layers          & UniformInt[1, 8] \\
Layer size                & UniformInt[64, 512] \\
Hidden dropout            & Uniform[0, 0.5] \\
Cross dropout             & Uniform[0, 0.5] \\
Learning rate             & LogUniform[1e-5, 1e-2] \\
Weight decay              & LogUniform[1e-6, 1e-3] \\
Category embedding size   & UniformInt[64, 512] \\
\midrule
\# Iterations             & 100 \\
\bottomrule
\end{tabular}
\caption{DCN V2 hyperparameter search space.}
\label{table:dcnv2_hyperparam}
\end{table}

\begin{table}[H]
\centering
\begin{tabular}{l|l}
\toprule
\textbf{Parameter} & \textbf{Distribution} \\
\midrule
\# Layers                & UniformInt[1, 6] \\
Feature embedding size   & UniformInt[8, 64] \\
Residual dropout         & Uniform[0.0, 0.2] \\
Attention dropout        & Uniform[0.0, 0.5] \\
Learning rate            & LogUniform[1e-5, 1e-3] \\
Weight decay             & LogUniform[1e-6, 1e-3] \\
\midrule
\# Iterations           & 100 \\
\bottomrule
\end{tabular}
\caption{AutoInt hyperparameter search space.}
\label{table:autoint_hyperparam}
\end{table}

\begin{table}[H]
\centering
\begin{tabular}{l|l}
\toprule
\textbf{Parameter} & \textbf{Distribution} \\
\midrule
\# Layers                & UniformInt[1, 4] \\
Feature embedding size   & UniformInt[64, 512] \\
Residual dropout         & Uniform[0, 0.2] \\
Attention dropout        & Uniform[0, 0.5] \\
FFN dropout              & Uniform[0, 0.5] \\
FFN factor               & Uniform[2/3, 8/3] \\
Learning rate            & LogUniform[1e-5, 1e-3] \\
Weight decay             & LogUniform[1e-6, 1e-3] \\
\midrule
\# Iterations           & 100 \\
\bottomrule
\end{tabular}
\caption{FT-Transformer hyperparameter search space.}
\label{table:fttransformer_hyperparam}
\end{table}

\begin{table}[H]
\centering
\begin{tabular}{l|l}
\toprule
\textbf{Parameter} & \textbf{Distribution} \\
\midrule
\# Layers                & UniformInt[1, 4] \\
Feature embedding size   & UniformInt[64, 512] \\
Residual dropout         & Uniform[0, 0.2] \\
Attention dropout        & Uniform[0, 0.5] \\
FFN dropout              & Uniform[0, 0.5] \\
FFN factor               & Uniform[2/3, 8/3] \\
Learning rate            & LogUniform[1e-5, 1e-3] \\
Weight decay             & LogUniform[1e-6, 1e-3] \\
\midrule
Min samples leaf        & UniformInt[1, 128] \\
Min impurity decrease   & LogUniform[1e-9, 0.01] \\
\midrule
\# Iterations           & 100 \\
\bottomrule
\end{tabular}
\caption{SG-XDEAT hyperparameter search space.}
\label{table:SG-XDEAT_hyperparam}
\end{table}

\section{Appendix C}
\subsection{Results for all algorithms on all datasets}
Table~\ref{table:main_results_2} summarizes the performance of various benchmark models across five datasets. To assess the statistical significance of performance differences, we apply the one-sided Wilcoxon test~\cite{wilcoxon1945individual} with a significance level of $\alpha = 0.05$ with Bonferroni correction. This provides a rigorous measure of whether improvements are consistent and not due to random variation.

\begin{table*}[t]
\centering
\begin{tabular}{l|ccccc}
\toprule
\textbf{Datasets} 
& \textbf{CA} 
& \textbf{GE} 
& \textbf{AD} 
& \textbf{KD} 
& \textbf{HI} \\
\cmidrule(lr){2-2} \cmidrule(lr){3-6}
\textbf{Metrics} 
& \multicolumn{1}{c}{RMSE\,$\downarrow$} 
& \multicolumn{4}{c}{Accuracy\,$\uparrow$} \\
\midrule
XGBoost        & \textbf{0.451±0.009} & \textbf{0.685±0.009} & \textbf{0.871±0.003} & \textbf{0.902±0.005} & 0.727±0.003 \\
\midrule
MLP            & 0.499±0.008 & 0.651±0.012 & 0.858±0.002 & 0.892±0.007 & 0.725±0.003 \\
Resnet         & 0.489±0.007 & 0.657±0.008 & 0.852±0.003 & 0.894±0.006 & \textbf{0.734±0.003} \\
DCN-V2         & 0.488±0.009 & 0.634±0.013 & 0.859±0.002 & 0.899±0.005 & 0.726±0.003 \\
AutoInt        & 0.490±0.009 & 0.602±0.015 & 0.859±0.002 & 0.898±0.004 & 0.726±0.003 \\
FT-Transformer & 0.472±0.009 & \textbf{0.677±0.009} & 0.861±0.002 & \textbf{0.903±0.005} & \textbf{0.732±0.002} \\
\midrule
SG-XDEAT       & \textbf{0.454±0.008} & 0.675±0.012 & \textbf{0.872±0.003} & \textbf{0.903±0.003} & \textbf{0.732±0.002} \\
\bottomrule
\end{tabular}
\caption{Comparison of performance across various benchmark models. Performance is reported as mean~$\pm$~standard deviation. The best result and those not statistically different from it ($p \geq 0.0083$) are shown in \textbf{bold}.}
\label{table:main_results_2}
\end{table*}

\subsection{Results for Architectural Ablation Analysis}
We present ablation results in Table~\ref{table:ablation_study_2} to evaluate the contribution of each architectural component in SG-XDEAT. Specifically, we compare the design (CD + CE) against two partial variants—CD-only and CE-only—as well as a baseline that directly concatenates feature embeddings (DFC) without modeling any structured attention. The full model consistently achieves comparable results, demonstrating the importance of modeling both types of dependencies.

\begin{table*}[t]
\centering
\begin{tabular}{l|ccccc}
\toprule
\textbf{Datasets} 
& \textbf{CA} 
& \textbf{GE} 
& \textbf{AD} 
& \textbf{KD} 
& \textbf{HI} \\
\cmidrule(lr){2-2} \cmidrule(lr){3-6} 
\textbf{Metrics} 
& \multicolumn{1}{c}{RMSE\,$\downarrow$} 
& \multicolumn{4}{c}{Accuracy\,$\uparrow$} \\
\midrule
DFC         & 0.480±0.010 & 0.649±0.015 & 0.869±0.003 & 0.902±0.005 & \textbf{0.732±0.003} \\
\midrule
CD          & 0.463±0.010 & \textbf{0.678±0.008} & 0.871±0.003 & \textbf{0.903±0.005} & \textbf{0.732±0.003} \\
CE          & 0.459±0.011 & 0.626±0.013 & \textbf{0.872±0.003} & 0.899±0.007 & 0.727±0.003 \\
\midrule
CD + CE     & \textbf{0.454±0.008} & 0.675±0.012 & \textbf{0.872±0.003} & \textbf{0.903±0.003} & \textbf{0.732±0.002} \\
\bottomrule
\end{tabular}
\caption{Ablation results for architectural components. Values are reported as mean ± standard deviation. \textbf{CD} = Cross-Dimension, \textbf{CE} = Cross-Encoding, \textbf{DFC} = Direct Feature Concatenation. \textbf{Best} results are highlighted.}
\label{table:ablation_study_2}
\end{table*}

\subsection{Results for Different Input Strategies}
Table~\ref{table: embeddings_2} reports the performance of models using either raw features or target-aware encodings exclusively. Across most datasets, neither stream consistently outperforms the other, suggesting that each captures distinct yet complementary information. However, the DFC method, which directly combines the raw and target-aware streams, fails to capture the dependencies between them effectively. These results motivate the dual-stream design adopted in SG-XDEAT, which aims to integrate both perspectives more effectively.

\begin{table*}[t]
\centering
\begin{tabular}{l|ccccc}
\toprule
\textbf{Datasets} 
& \textbf{CA} 
& \textbf{GE} 
& \textbf{AD} 
& \textbf{KD} 
& \textbf{HI} \\
\cmidrule(lr){2-2} \cmidrule(lr){3-6} 
\textbf{Metrics} 
& \multicolumn{1}{c}{RMSE\,$\downarrow$} 
& \multicolumn{4}{c}{Accuracy\,$\uparrow$} \\
\midrule
Raw             & 0.483±0.008 & \underline{0.665±0.013} & 0.859±0.002 & \underline{\textbf{0.903±0.005}} & \textbf{0.732±0.003} \\
Targeted    & \underline{0.479±0.011} & 0.655±0.015 & \underline{0.863±0.005} & 0.888±0.007 & \textbf{0.732±0.002} \\
\midrule
DFC             & 0.480$\pm$0.010 & 0.649$\pm$0.015 & 0.869$\pm$0.003 & 0.902$\pm$0.005 & \textbf{0.732$\pm$0.003} \\
\midrule
CD + CE         & \textbf{0.454$\pm$0.008} & \textbf{0.675$\pm$0.012} & \textbf{0.872$\pm$0.003} & \textbf{0.903$\pm$0.003} & \textbf{0.732$\pm$0.002} \\
\bottomrule
\end{tabular}
\caption{Performance comparison between different input strategies. The \textit{Raw} setting uses original features only, while the \textit{Targeted} variant incorporates label-dependent encodings. The DFC and CD+CE methods are described in Figure~2 (Main Text). Values are reported as mean~$\pm$~standard deviation. The overall best results are shown in \textbf{bold}, while \underline{underlined} values indicate the better performance between Raw and Targeted for each dataset.}
\label{table: embeddings_2}
\end{table*}

\subsection{Setup for Adaptive Sparse Self-Attention}
This experiment aims to evaluate whether incorporating Adaptive Sparse Self-Attention (ASSA) into a pre-norm Transformer improves model robustness in the presence of noisy or redundant features. To this end, we construct a controlled synthetic regression benchmark where the degree of feature redundancy is systematically varied. Table~\ref{table:ASSA} reports the test RMSE for pre-norm Transformers with and without ASSA across different values of $\rho$. The results demonstrate that models equipped with ASSA consistently achieve lower RMSEs confirming its effectiveness in suppressing noisy features and enhancing predictive robustness.

\subsubsection{Synthetic Benchmark Construction}
We construct a synthetic regression benchmark to test whether incorporating Adaptive Sparse Self-Attention (ASSA) into a pre-norm Transformer improves performance compared to using standard softmax attention only. Each input is a 100-dimensional vector, with only a fraction $\rho \in {0.5, 0.6, \dots, 1.0}$ containing useful signal. For each $\rho$ and random seed (10 in total), we generate 64,000 training, 16,000 validation, and 20,000 test samples. Targets are computed using a fixed, randomly initialized 4-layer MLP with ReLU activations between layers, which processes only the informative features. The same MLP is used across all splits for a given seed and $\rho$. Target values are standardized to zero mean and unit variance.

This setup produces 60 datasets (10 seeds × 6 $\rho$ values), allowing for fine-grained evaluation of model behavior under increasing feature sparsity. It provides a clear testbed for assessing whether ASSA-equipped Transformers are more robust to noisy or redundant features compared to their softmax-based counterparts.

\subsubsection{Implementation Details}
In this setup, we do not apply any target-aware encoding. Instead, raw input features are directly tokenized and fed into a standard Transformer backbone. The full configuration is summarized in Table~\ref{table:Transformer_default}. Training is conducted for up to 200 epochs with a batch size of 512. Early stopping is employed based on validation performance, using a patience of 10 epochs.

\begin{table}[H]
\centering
\begin{tabular}{l|l}
\toprule
\textbf{Parameter} & \textbf{Setup} \\
\midrule
\# Layers               & 3 \\
\# Heads                & 8 \\
Feature embedding size  & 192 \\
Feature hidden size     & 256 \\
Residual dropout        & 0.0 \\
FFN dropout             & 0.1 \\
Attention dropout       & 0.2 \\
\midrule
( Optimizer, LR )          & (AdamW, 1e-3) \\
\bottomrule
\end{tabular}
\caption{Transformer Configuration for ASSA Experiments}
\label{table:Transformer_default}
\end{table}

\begin{table*}[t]
\centering
\begin{tabular}{l|cccccc}
\toprule
\textbf{Setting} 
& $\rho=0.5$ 
& $\rho=0.6$ 
& $\rho=0.7$ 
& $\rho=0.8$ 
& $\rho=0.9$ 
& $\rho=1.0$ \\
\midrule
W/ ASSA     & \textbf{0.498±0.032} & \textbf{0.517±0.032} & \textbf{0.529±0.055} & \textbf{0.565±0.033} & \textbf{0.593±0.031} & \textbf{0.616±0.055} \\
W/O ASSA    & 0.523±0.024 & 0.556±0.034 & 0.584±0.060 & 0.617±0.030 & 0.646±0.028 & 0.651±0.036 \\
\bottomrule
\end{tabular}
\caption{Regression performance (RMSE) on synthetic datasets with varying proportions of informative features $\rho$. A lower $\rho$ implies a higher proportion of irrelevant (noisy) features. Values are reported as mean ± standard deviation.}
\label{table:ASSA}
\end{table*}

\end{document}